\documentclass[letterpaper, 10 pt, conference]{ieeeconf}  

\usepackage{times}
\usepackage{epsfig}
\usepackage{graphicx}
\usepackage{amsmath}
\usepackage{amssymb}
\usepackage[pagebackref=true,breaklinks=true,colorlinks,bookmarks=false]{hyperref}
\usepackage[linesnumbered,ruled,vlined]{algorithm2e}

\makeatletter
\def\BState{\State\hskip-\ALG@thistlm}
\makeatother

\def\icarus/{{\sc Icarus}}

\usepackage{footnote}

\usepackage{xspace}
\newcommand{\etal}[0]{{\em et al.~}}
\newcommand{\eg}[0]{{\em e.g.,~}}
\newcommand{\ie}[0]{{\em i.e.,~}}
\newcommand{\etc}[0]{{\em etc.\xspace}}

\usepackage{tikz}
\usetikzlibrary{calc, math}
\usepackage{etoolbox}
\makeatletter
\patchcmd{\@verbatim}
  {\verbatim@font}
  {\verbatim@font\small}
  {}{}
\makeatother
\IEEEoverridecommandlockouts                              

\title{\LARGE \bf
Improving Object Permanence using Agent Actions and Reasoning
}

\author{Ying Siu Liang, Chen Zhang, Dongkyu Choi and Kenneth Kwok$^{1}$
\thanks{$^{1}$All authors are with Social and Cognitive Computing at the Institute of High Performance Computing, Agency for Science, Technology and Research, Singapore
        {\tt\small \{liangys, zhang\_chen, choi\_dongkyu, kenkwok\}@ihpc.a-star.edu.sg}}%
}

\begin{document}

\maketitle
\thispagestyle{empty}
\pagestyle{empty}

\begin{abstract}

Object permanence in psychology means knowing that objects still exist even if they are no longer visible. It is a crucial concept for robots to operate autonomously in uncontrolled environments. Existing approaches learn object permanence from low-level perception, but perform poorly on more complex scenarios, like when objects are \textit{contained} and \textit{carried} by others. Knowledge about manipulation actions performed on an object prior to its disappearance allows us to reason about its location, \eg that the object has been placed in a carrier.
In this paper we argue that object permanence can be improved when the robot uses knowledge about executed actions and describe an approach to infer hidden object states from agent actions. 
We show that considering agent actions not only improves rule-based reasoning models but also purely neural approaches, showing its general applicability.
Then, we conduct quantitative experiments on a snitch localization task using a dataset of 1,371 synthesized videos, where we compare the performance of different object permanence models with and without action annotations. 
We demonstrate that models with action annotations can significantly increase performance of both neural and rule-based approaches. 
Finally, we evaluate the usability of our approach in real-world applications by conducting qualitative experiments with two Universal Robots (UR5 and UR16e) in both lab and industrial settings. The robots complete benchmark tasks for a gearbox assembly and demonstrate the object permanence capabilities with real sensor data in an industrial environment.

\end{abstract}

\section{INTRODUCTION}
\label{sec:intro}
Object permanence is a fundamental concept studied in developmental psychology. It describes the capacity to understand that objects continue to exist even when they become unobservable.
For infants, it has been identified as a precursor to language acquisition~\cite{tomasello1984cognitive} and the development of searching behaviors~\cite{ahsen2017object}.
Modeling this behavior in robotic agents generally addresses the simpler instance of object permanence, referred to as \textit{visible displacements}, which can be treated as a low-level reasoning problem where objects are tracked based on perceived pixels, associating bounded regions of subsequent frames and comparing their color, shape, or location~\cite{michel2004motion,papadourakis2010multiple,roy2004mental}.

\begin{figure}[tb]
	\centering	
 	\includegraphics[width=0.9\linewidth]{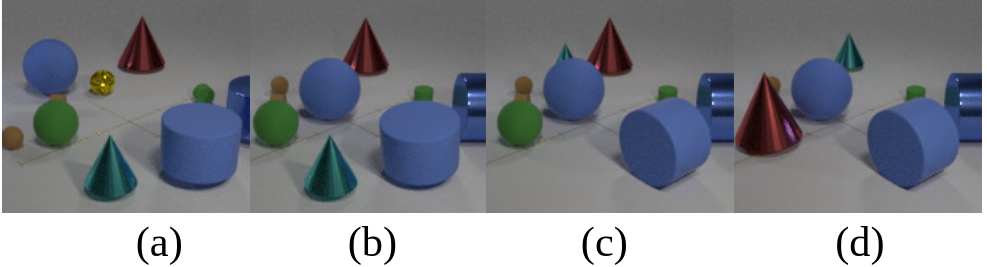}%
 	\vskip -0.1in
 	\caption{\textit{Invisible displacement} in the snitch localization task \cite{shamsian2020learning}: the snitch becomes occluded by the blue ball (b), the green cone is moved behind the blue ball (c), but it is impossible to tell if it contains the snitch (d).}
	\label{fig:intro-seq}
	\vskip -0.1in
\end{figure}

Far less work addresses the more complex level of object permanence, known as \textit{invisible displacements}~\cite{Singer2018}, that deals with hidden objects in motion (\ie that are \textit{carried}). 
Deep learning approaches that address this problem generally require large amounts of data to train a model and mostly rely on visual perception~\cite{shamsian2020learning}. 
While object motion trajectories can be used to predict future states and possible occlusions~\cite{papadourakis2010multiple}, some predictions cannot be made from visual information alone. 
For example, in Fig.~\ref{fig:intro-seq}, the snitch (a small golden sphere) becomes occluded by the blue ball, and the green cone is moved behind the ball. Here, there are two possibilities: the green cone either contains the snitch or not. Thus, when the green cone moves, the snitch might move with the green cone (while contained) or it might simply stay behind the blue ball. Disambiguating this situation requires high-level reasoning using knowledge about the action performed, namely, if the green cone was indeed placed on top of the snitch when moving behind the blue ball.
In robotics, \textit{perceptual anchoring} is commonly used to track objects in the world by building and maintaining a correspondence between sensor data and the symbols denoting to the same physical objects through high-level reasoning~\cite{coradeschi2003introduction}.

\begin{figure}[t]
	\centering	
 	\includegraphics[width=1\linewidth]{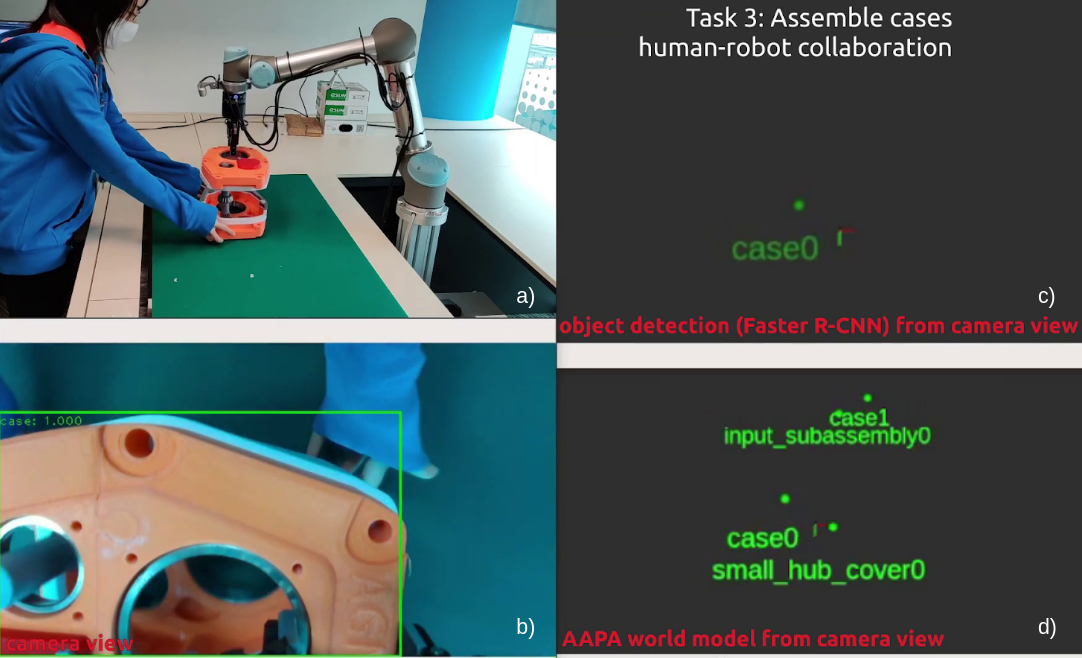}%
 	\vskip -0.1in
 	\caption{Collaborative assembly task in the lab setting: robot's camera view is obstructed by the held object (b), surrounding objects are lost by the object detection (c), but maintained with our action-aware perceptual anchoring model (d). Only tracked object names are visualized for simplicity.}
	\label{fig:demo-overview}
	\vskip -0.1in
\end{figure}

In recent work~\cite{icra2021}, we proposed an Action-Aware Perceptual Anchoring model (AAPA) that takes a rule-based approach allowing robots to reason about object permanence from agent actions. The model introduces inductive biases to handle situations where objects get occluded, move out of view, or flicker on and off due to noisy detection.
Furthermore, we defined an \textit{attachment} relation to be the state between two objects where they are physically bound to move together, generally resulting from an action such as grasping, inserting or screwing on.

In this paper, we take one step further and argue that using agent actions can improve object permanence capabilities in both rule-based approaches like AAPA, as well as in purely neural models like OPNet~\cite{shamsian2020learning}. 
First, we talk about related work that addresses the object permanence problem (Sec.~\ref{sec:related-work}). Then we describe our approach how agent actions can be used to infer attachment relations and hidden object states (Sec.~\ref{sec:approach}). Following this, we talk about how inferred attachment relations can be used by neural models (OPNet+AA); and give an overview of our action-aware rule-based model (AAPA) and how it uses agent actions. 
We present quantitative experiments to compare performances of existing baselines to models that consider agent actions, as well as different model variants of AAPA (Sec.~\ref{sec:experiments}). We use the LA-CATER dataset~\cite{shamsian2020learning} and their snitch localization task to measure mean Intersection over Union (IoU) and Euclidean distance ($L^2$-distance).
For a qualitative evaluation, we present experiments in a real-world application where we use AAPA on two Universal Robots (UR5 and UR16e) to infer hidden object states from their own executed actions (Sec.~\ref{sec:sys-eval}). We show that AAPA can easily be used for different robots in a lab and industrial setting using a manufacturing domain. Our experiments demonstrate that agent action information can significantly improve the object tracking performance for more complex tasks such as invisible displacements. 
Finally, we conclude with a discussion on obtained results and future research directions (Sec.~\ref{sec:conclusion}).
\section{RELATED WORK}\label{sec:related-work}
\vspace{-0.05in}
The problem of object permanence is commonly addressed in the context of visual object tracking and long-term occlusion.
Reid \cite{reid1979algorithm} first proposed an algorithm for tracking multiple targets where a Kalman Filter is used to estimate target states. 
Temporally long object occlusions have been addressed by labeling spatially close objects as \textit{occluders}~\cite{papadourakis2010multiple}.
Siamese networks have become more popular and achieved state-of-the-art results in visual object tracking and can handle full occlusions and objects going out of view~\cite{dong2018triplet,he2018twofold,li2018high,tao2016siamese,wang2019fast,zhu2018distractor}.
Lang \etal \cite{lang2018deep} trained a CNN to learn a sense of agency and object permanence, where the robot predicts objects that are occluded by its own joints.
Models for inferring containment relations have been proposed previously~\cite{liang2016inferring,ullman2019model}, which can be used to recover incomplete object trajectories~\cite{liang2018tracking}. 
Recent work by Shamsian \etal \cite{shamsian2020learning} presented the OPNet architecture, a neural model that outperforms baseline models on the LA-CATER dataset on carried tasks. It is trained purely on the video data and does not consider action annotations.
Object permanence is also addressed in developmental robotics using perceptual anchoring to allow robots to reason about the environment~\cite{milliez2014framework,roy2004mental,blodow2010perception,heyer2012intelligent}. 
These approaches use semantics and high-level reasoning to maintain anchors for objects in the world but do not make use of agent actions to predict object attachments and hidden states.

\begin{figure}[thbp]
	\centering	
 	\includegraphics[width=1\linewidth]{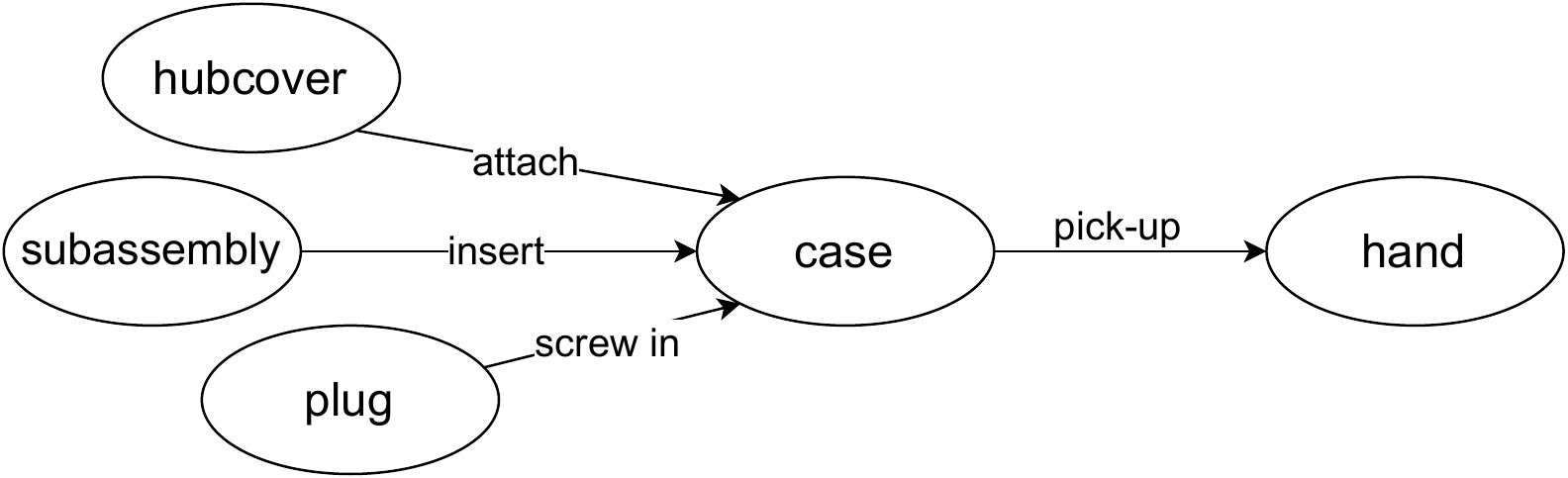}%
 	\vskip -0.1in
 	\caption{Example of an attachment hierarchy generated by agent actions.}
	\label{fig:attachment-hierarchy}
	\vskip -0.1in
\end{figure}
	\vskip -0.05in
\section{ACTIONS and ATTACHMENT HIERARCHY}\label{sec:approach}
Every action executed by an agent can have \textit{effects} on the world state.
For example, if we know that an object has been picked up, we can infer that the hand is now holding the object. From the robot's point of view, this is equivalent to having executed a {\tt pick-up} action and inferring that a new relation, {\tt holding}, is true for the robot's hand and that object in the subsequent world states until a counteraction like {\tt put-down} is executed for the same combination of objects.
In~\cite{icra2021}, we defined a higher-order semantic relation, {\tt attached}($p_i,p_j$) between two objects, where a child object $p_i$ is \textit{attached} to a parent object $p_j$ and therefore $p_i$ is physically constrained to move together with $p_j$.
Given an agent action, $a_t(p_i, p_j)$, at a given time step $t$, we can define an attachment relation {\tt attached}($p_i,p_j$) that holds for time steps $t'$, such that $t \leq t'<T$ and there exists an action, $a_T(p_i, p_j)$, at time step $T$ that counteracts $a_t$.
For example, given
{\tt pick-up$_0$}(hand,obj) and {\tt put-down$_9$}(hand,obj), we can infer that {\tt attached}(obj,hand) will hold for time steps $t'\in[0,8]$.
Furthermore, we assume that this is a one-to-many relation, where an object can be attached to only one parent at any point in time, but one parent can have multiple children.
In more complex scenarios, such as those dealing with a gearbox assembly, multiple consecutive actions can generate an attachment hierarchy $\mathcal{H}$ like shown in Fig.~\ref{fig:attachment-hierarchy} that involves \{(hubcover, case), (subassembly, case), (plug, case), (case, hand)\}.
Thus, if the highest-order parent object is visible (\ie hand), we assume that their children (\ie case), as well as the children of those (\ie hubcover, subassembly, plug) still physically exist in the scene, even if they are not perceived anymore, and can maintain their symbol in the robot's world model.

Given a set of attach and detach actions $\mathcal{AD}$ and a list of executed actions $\mathcal{A}$, we can generate the attachment hierarchy for each time step $t$ (Algorithm~\ref{alg}). 
$\mathcal{AD}$ can be domain-dependent (\eg manufacturing) and is defined manually as learning this is beyond the scope of this paper.
In the following, we will describe how our attachment hierarchy can be used in a purely neural model (Sec.~\ref{subsec:opnet-aa}) as well as in our rule-based model AAPA~\cite{icra2021} (Sec.~\ref{subsec:aapa}).

\begin{algorithm}
 \KwData{$\mathcal{AD} = \{(a,d)\}$: attach-detach actions\;
 $\mathcal{A} = \{a_t\}$: executed actions at time step $t$
}
 \KwResult{$\mathcal{H} = \{(c,p)\}$: attachment hierarchy}
$\mathcal{H} \gets \emptyset$\;
 \For{each $a_t$ in $\mathcal{A}$}{
  $c \gets child(a)$;
  $p \gets parent(a)$\;
  \uIf{$a_t$ is attach in $\mathcal{AD}$}{
   $\mathcal{H} \gets (c,p)$\;
   }{
   \uElseIf{$a_t$ is detach in $\mathcal{AD} \And (c,p)$ in $\mathcal{H}$}{
   $\mathcal{H} \gets \mathcal{H}\setminus(c,p)$ \;}
    }
    }
 \caption{Generate attachment hierarchy}
 \label{alg}
\end{algorithm}

\vskip -0.1in
\subsection{Neural models with Action Annotation}
\label{subsec:opnet-aa}
For neural models, there are different ways to incorporate the attachment hierarchy.
We can include them as a simple numeric feature \eg a 3-dimensional feature describing action and attachment relation $(action,child,parent)$ and train the model with this additional input.
A more efficient way is to provide the model with explicit guidance, \ie generate a weight matrix from the attachment hierarchy and incorporate it into the model.

\begin{figure}[thbp]
	\centering	
 	\includegraphics[width=1\linewidth]{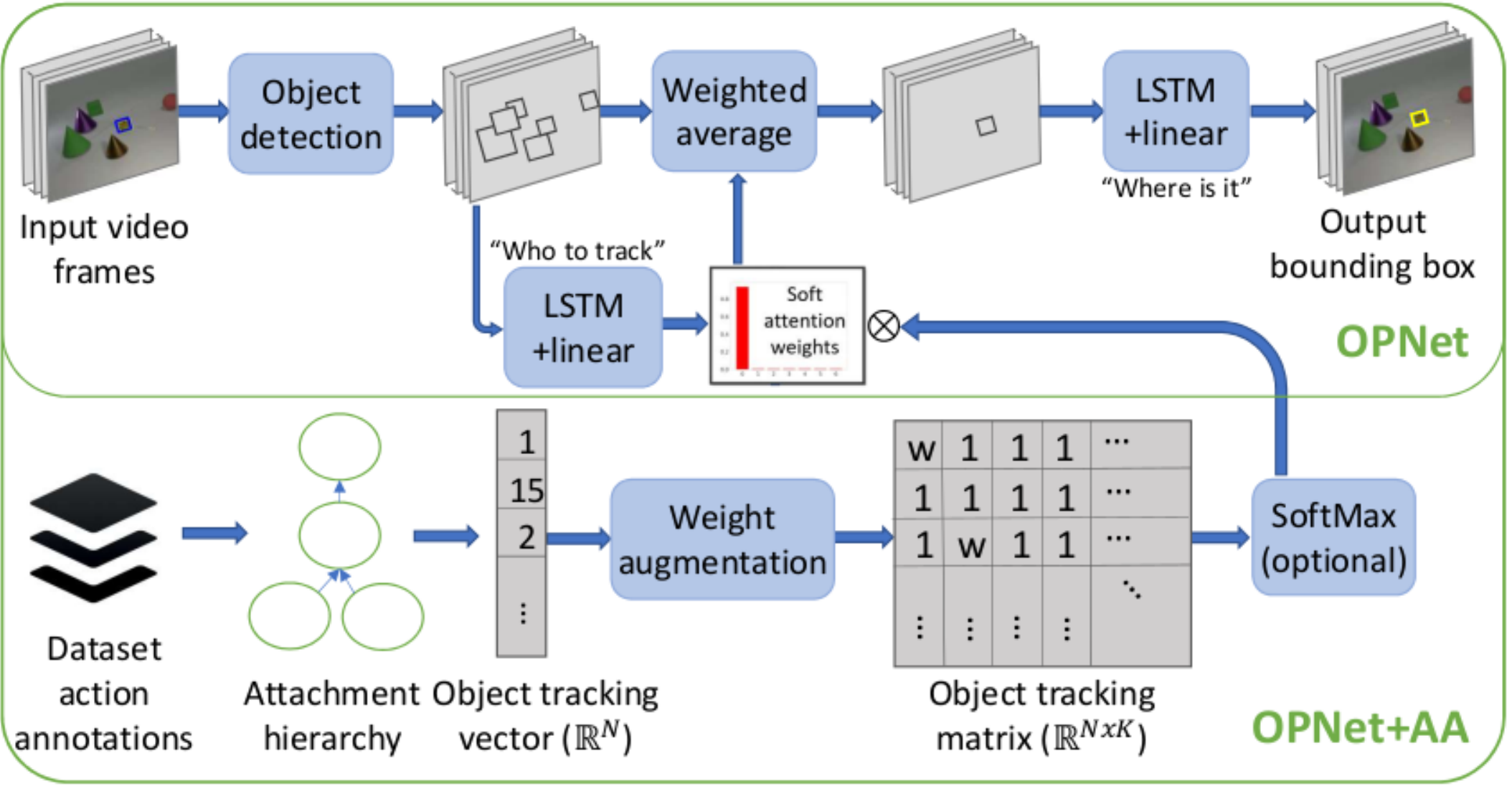}%
 	\vskip -0.1in
 	\caption{OPNet+AA: agent action information injected into OPNet \cite{shamsian2020learning}}
	\label{fig:opnet-aa}
	\vskip -0.1in
\end{figure}

For this, we first create an object tracking vector $v\in\mathbb{N}^{N}$, where $N$ is the total number of frames in a training video. Given the object positions from ground-truth scene annotations, 
we use the attachment hierarchy to populate $v$ such that it contains the ID of the object to track for each frame: this could be the object itself (if it is visible in that frame), or the ID of a parent in the attachment hierarchy. If multiple actions have been performed, the ID is set to the parent at the highest level of the attachment hierarchy.
We then generate a weight matrix $W\in\mathbb{R}^{N\times K}$ from $v$, where $N$ is the total frame count and $K$ is the maximum object count, and use a weight parameter $w$ to control the impact of the explicit guidance:
\begin{enumerate}
    \item Initialize $W$ with elements all equal to 1;
    \item for each frame $t$, look up the object to track ($v_t$) and multiply the corresponding element in $W_t$ with $w$;
    \item (optional) use SoftMax on the matrix to obtain a normalized weight.
\end{enumerate}
Theoretically, this augmented attention incorporates explicit guidance on the object to track and we can vary its strength by changing $w$. We will show how this is augmentation is implemented the neural model OPNet in Sec.~\ref{sec:experiments} (Fig.~\ref{fig:opnet-aa}).
\vskip -0.1in

\begin{figure}[hbp]
	\centering	
 	\includegraphics[width=\linewidth]{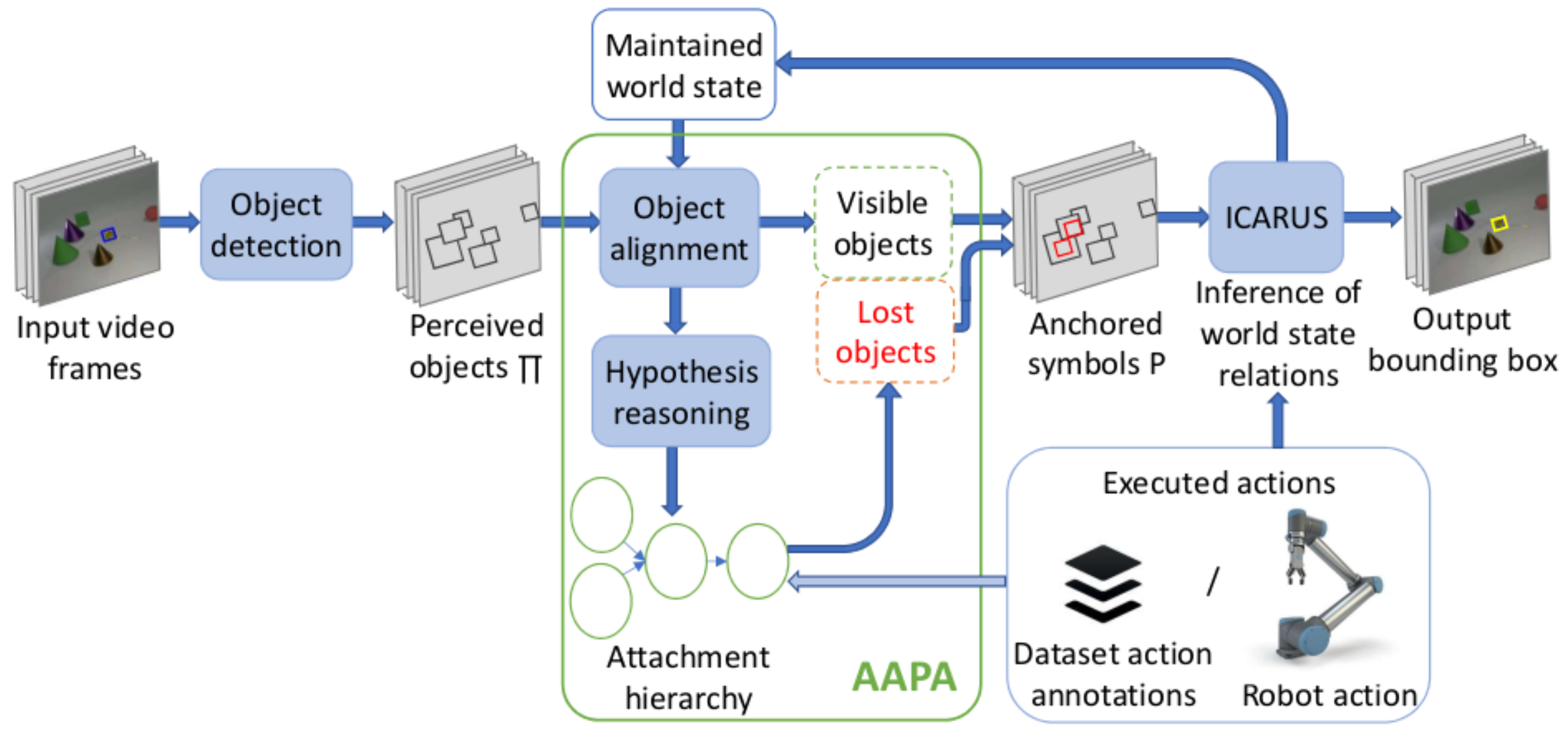}
 	\vskip -0.1in
 	\caption{Overview of the end-to-end perceptual anchoring process: an object detection module segments a raw camera image into objects $\Pi$ and passes them to AAPA integrated into a cognitive architecture, \icarus/. AAPA performs object alignment and rule-based reasoning to maintain anchors in the symbol system $\mathcal{P}$. \icarus/ then infers its current beliefs based on the world states and executes relevant actions. The world state $\mathcal{S}$ and executed actions $\mathcal{A}$ are passed as input to AAPA during the subsequent cycle.}
	\label{fig:model-overview}
	\vskip -0.1in
\end{figure}

\subsection{Action-Aware Perceptual Anchoring (AAPA)}
\label{subsec:aapa}
In~\cite{icra2021}, we proposed an action-aware perceptual anchoring model (AAPA), a rule-based model that uses high-level reasoning to predict hidden object states. We implemented AAPA as part of the cognitive architecture \icarus/~\cite{Choi2018}, a framework for modeling human cognition that includes world state inference (from a given world model) and high-level task planning. We made the assumption that object state changes are observable, gradual and consistent:
\begin{enumerate}[leftmargin=*]
    \item \textbf{Observable:} Objects do not move by themselves and all movements are caused by some agent actions. As long as no agent moves them, objects remain in the same position as the last time they were observed until they are perceived again.
    \item \textbf{Gradual:} Objects do not `teleport'. Therefore, objects detected in successive frames have similar attributes (like position and size). 
    \item \textbf{Consistent:} Objects that exist in the world will be detected consistently over time. Hence, objects are assumed to exist (or not exist) in the symbolic world model only if they are perceived (or missing) for several consecutive cycles. 
\end{enumerate}
These assumptions allowed us to make reliable inferences on the positions of objects that are no longer detected by the perceptual system but are physically still present in the scene. 
In the following we give a brief overview of the end-to-end pipeline (Fig.~\ref{fig:model-overview}), which consists of three main steps:
\begin{enumerate}[leftmargin=*,noitemsep]
    \item[a)] {\textbf{Object detection:} At each time step $t$, an object detector segments the raw image data into a set of objects, $\Pi_t = \{\pi_i : i = 1 \dots I \}$ and passes it to AAPA. Each object $\pi_i$ is associated with a set of attributes, $\Phi_i = \{\phi_i^{type},\phi_i^{pos},\phi_i^{size}\}$, describing the object's type and its detected bounding box position and dimension.}
    \item[b)] {\textbf{AAPA} first solves the \textit{assignment problem} using Munkres~\cite{munkres1957algorithms} to find the optimal object alignment between the newly perceived objects $\Pi_t$ and the previously maintained world state $\mathcal{S}_{t-1}$. 
    It uses the object attributes $\Phi_i$ to compute a cost matrix, where object alignment dissimilarities are capped at a maximum threshold parameter, $\tau$ (which can be adjusted according to the reliability of the object detection output). Objects that are aligned successfully are considered \textit{visible} and added to the set of anchored symbols $\mathcal{P}_t$.
    AAPA then performs hypothesis reasoning, where it tries to anchor objects that were not aligned, or \textit{lost}, and adds them to $\mathcal{P}_t$ (discussed below).
    }
    \item[c)] {\textbf{ICARUS} takes the anchored symbols $\mathcal{P}_t$, the last executed actions $\mathcal{A}_{t-1}$ and infers its belief state of the world, $\mathcal{S}_t(\supseteq\mathcal{P}_{t})$, which includes the anchored objects $\mathcal{P}_t$ as well as object relations describing how objects relate to one another. These relations can be inferred from object positions alone, \eg 
    {\tt left-of}(plug,case), or from $\mathcal{A}_{t-1}$. As described in Algorithm \ref{alg}, we check if  $\mathcal{A}_{t-1}$ is in the set of attach-detach actions $\mathcal{AD}$ and infer object attachments accordingly (\eg {\tt insert}(plug,case) $\mapsto$ {\tt attached}(plug,case)). Thus, all {\tt attached} relations in $\mathcal{S}_t$ form the attachment hierarchy $\mathcal{H}$.
    Based on this belief state and a given goal, \icarus/ performs high-level task planning for the next executable action $\mathcal{A}_t$. }
\end{enumerate}
Due to limited space we only provide details on AAPA's hypothesis reasoning component that uses agent actions to generate an internal attachment hierarchy.
More details on the model and other components can be found in~\cite{icra2021}.

\subsubsection{Hypothesis reasoning with attachment hierarchy} 
\label{sssec:hypothesis}
The hypothesis reasoning component considers \textit{lost} objects, \ie previously maintained anchors that have not been aligned successfully in the object alignment process (${p}_j\in{\mathcal{P}}_{t-1}$, ${p}_j\notin\mathcal{P}_{t}$). 
Based on the assumptions that object state changes are observable, gradual, and consistent, this component reasons about occlusions, potential detection errors and other state changes to decide if their anchors should be maintained in $\mathcal{P}_t$. 
To consider agent actions, we added a reasoning process that checks for attachment relations in the previously maintained world state $\mathcal{S}_{t-1}$ and update hidden object states in the current anchored set $\mathcal{P}_t$. It recursively traverses the attachment hierarchy until it finds a parent that has been anchored already.
More formally, for each symbol $p_j \in \mathcal{P}_{t-1}$ that has not been anchored ($p_j \notin \mathcal{P}_{t}$), if there exists a symbol $p_k \in \mathcal{P}_{t}$ and a relation {\tt attached}$(p_j, p_k) \in \mathcal{S}_{t-1}$, then the child $p_j$ is considered to be \textit{attached} to the parent $p_k$ and its position is updated to follow that of $p_k$. 
Thus, the anchor for $p_j$ is added to $\mathcal{P}_t$.
Note that $p_k$ does not need to be visible itself, but might have been anchored due to another parent $p_n$.
In summary, an object anchor that is not visible is updated and maintained, if it has a higher-order parent that is already anchored. Conversely, if an object has been anchored, all of its children are anchored as well. 
\section{QUANTITATIVE EXPERIMENTS}
\label{sec:experiments}
For a quantitative evaluation we present experiments comparing performances of existing baselines that do not use agent actions to models that do. We implemented OPNet+AA, an augmented version of the purely neural model OPNet~\cite{shamsian2020learning} and different variants of our rule-based model AAPA with and without the action annotation component: 
\begin{enumerate}[leftmargin=*,noitemsep]
                \item \textbf{OPNet}~\cite{shamsian2020learning}: a purely neural model that consists of two LSTMs acting as reasoning modules for inferring the object to track (if it is occluded or carried) and its location. The model uses an attention mask that focuses on the object that covers the target in a frame, but is not explicitly guided.
                \item \textbf{OPNet+AA}: OPNet with agent action information, where the attention mask receives explicit guidance generated from ground-truth annotations and attachment relations, otherwise trained using the same setup as OPNet.
                \item \textbf{PA}: our perceptual anchoring model AAPA but without considering attachment relations.
                \item \textbf{AAPA-3k, AAPA-6k5, AAPA-10k}: variants of AAPA with different object alignment thresholds $\tau$ = 3000, 6500, 10000 respectively.
\end{enumerate}
In \cite{icra2021} we compared overall model performances between OPNet, AAPA-6k5, and a heuristic model, that switches from tracking the snitch to tracking the object closest to its last known location. In this work, we were interested in comparing a) OPNet vs OPNet+AA, b) PA vs AAPA, c) different AAPA variants, and d) OPNet+AA vs AAPA.
In the following we provide details on the dataset used (\ref{subsec:dataset}), 
the experiment setup (\ref{subsec:quan-setup}) and metrics (\ref{subsec:metrics}) and discuss experimental results (\ref{subsec:quant-results}).

\subsection{Dataset}
\label{subsec:dataset}

We run experiments using the LA-CATER (Location Annotations) dataset~\cite{shamsian2020learning} on a snitch localization task.
The dataset consists of 13,998 videos split into train (9,300), validation (3,327) and test (1,371) datasets (Table~\ref{tab:la-cater}).
The authors divided the dataset into four object permanence tasks with increasing complexity (visible, occluded, contained, carried). LA-CATER was generated from the CATER dataset~\cite{girdhar2020cater} but includes frame-level annotations for the four subtasks and object ground-truth locations for the entire video sequences. It enables better comparisons between different models on more complex object permanence tasks. 
Each 300-frame video (320x240 pixel) is synthetically rendered with standard 3D objects (cube, cylinder, sphere, inverted cone) of different size (small, medium, large), material (metal, rubber) and color (eight colors). Action annotations describe a set of basic actions (rotate, slide, contain, pick \& place), which can be considered as agent actions performed by an invisible hand. Each video contains exactly one snitch and 5-15 unique objects in total. The camera does not move and objects stay in the scene for the entire duration.
The \textit{contain} action signifies a \textit{pick \& place}, which results in an inverted cone containing another object.
Each video contains a small golden sphere, referred to as the `snitch', that is used as the target in the localization task.

\begin{table}[thbp]
\caption{Distribution of LA-CATER~\cite{shamsian2020learning} dataset}
	\vskip -0.2in
\label{tab:la-cater}
\begin{center}
\begin{tabular}{l||ccccc}
& \# & visible & occluded & contained & carried \\
\hline
Train & 9,300            & 63.00\% & 3.03\%   & 29.43\%   & 4.54\%  \\
Dev   & 3,327            & 63.27\% & 2.89\%   & 29.19\%   & 4.65\%  \\
Test  & 1,371            & 64.13\% & 3.07\%   & 28.56\%   & 4.24\%  \\
\end{tabular}
\end{center}
\vskip -0.2in
\end{table}

\vskip -0.1in

\subsection{Experiment setup}\label{subsec:quan-setup}
We used the code 
provided by the authors~\cite{shamsian2020learning} to train and test the baseline models.
In OPNet, there is a ``who to track'' module responsible for understanding which object is currently covering the target. It consists of a single LSTM layer with a hidden dimension of 256 neurons and a linear projection matrix. It projects the LSTM output to $K$ neurons, then applies a SoftMax layer to compute the attention mask.
For OPNet+AA, we provide this attention mask with explicit guidance and generated the attachment hierarchy using $\mathcal{AD}=\{($\textit{contain},\textit{pick\&place})\} as described in Algorithm \ref{alg} and Sec.~\ref{subsec:opnet-aa} (Fig.~\ref{fig:opnet-aa}). 
For the rest of our experiments, we kept the setup of the original paper, including the model design and hyper-parameter setup, as well as training and validation data. 
We also experimented with different weights $w$, with and without SoftMax, where we obtained different performances. Due to limited space, we only include the model which yielded the best performance ($w=100$ no SoftMax). 

In addition to several variants of AAPA that use different object alignment thresholds $\tau$, we also included a model (PA) that does not consider object attachments during the hypothesis reasoning process. This serves as a baseline to evaluate the impact of agent actions.
All models including the AAPA variants use the same object detection input as OPNet. This input is received either from their trained Faster R-CNN object detection (OD) or ground-truth bounding boxes, \ie `perfect perception' (PP).
We used the action annotation provided by the dataset to infer attachment relations for both OPNet+AA and AAPA. 

\subsection{Metrics}\label{subsec:metrics}
To compare the models, we use the mean Intersection over Union (IoU) as well as the mean Euclidean distance ($L^2$-distance) of the bounding box centers.
Comparing the mean $L^2$-distance provides us with another measure of how reliable the models are, by comparing only the tracked object position and mitigating the effect of bounding box size errors. Furthermore, we only compute the metrics from the moment, when the snitch is first detected by the object detector, as any predictions before that would be impossible or random. 
The $L^2$-distance is given in pixels. As a reference, the video size is 360x240 pixel, the ground-truth size of the snitch ranges from 15x16 to 41x41 and of any object in the scene from 8x8 to 117x117.

\begin{table*}[ht]
  \centering
  \caption{Results comparing mean IoU (higher is better) and mean $L^2$-distances (lower is better) with standard errors of mean (SEM)}
   \vskip -0.15in
  \label{tab:model-comparison}
  \tikzmath{real \r; \r=2.1; }
  \begin{tikzpicture}[scale=1.0]
  
    \path (\r+.5, 0) node[right, scale=1.0] 
    {
      \tikzmath{real \r; \r=2.1; }
      \begin{tikzpicture}[scale=1.0]
        \path (\r+.5, 0) node[right, scale=1.0] {%
        \includegraphics[width=19.8em]{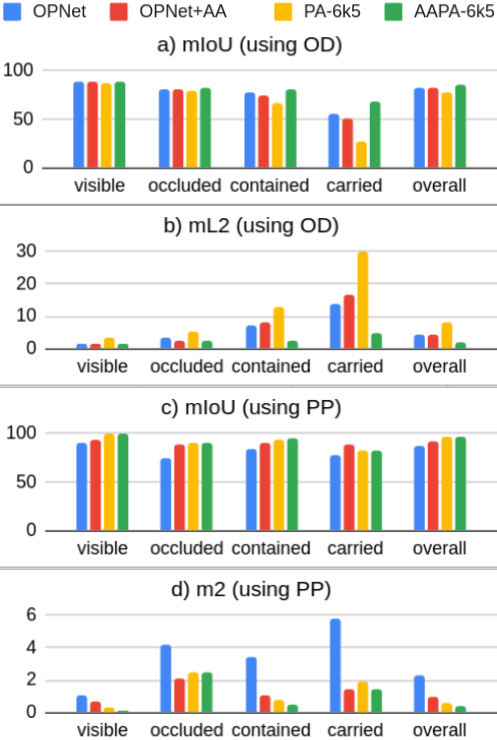}
        };
      \end{tikzpicture}
    };
    \path (\r+.6, 0) node[left, scale=1.0] {%
        \begin{tabular}{rccccc}
        \hline
        \multicolumn{6}{|c|}{\textbf{I. Using Object Detection (OD)}} \\
        \hline
        \textbf{a) mIoU} $\pm$SEM & Visible & Occluded & Contained & Carried & Overall \\
        \hline
        OPNet 	    & \textbf{88.98 $\pm$0.15}  &  80.19 $\pm$0.51  & 77.07 $\pm$0.61  &  56.04 $\pm$0.77  &  	82.35  $\pm$0.39  \\
        OPNet+AA	& 88.54 $\pm$0.19  &  81.44 $\pm$0.38  & 74.10 $\pm$0.62  &  50.75 $\pm$0.76  &  	81.54  $\pm$0.40  \\ 
        PA-6k5 	    & 86.80 $\pm$0.31  &  79.43 $\pm$0.55  & 67.14 $\pm$0.76  &  26.54 $\pm$0.43  &  	76.96  $\pm$0.53  \\
        AAPA-3k	    & 85.81 $\pm$0.37  &  78.74 $\pm$0.58  & 75.38 $\pm$0.57  &  54.57 $\pm$0.69  &  	80.48  $\pm$0.45  \\
        AAPA-6k5	& 88.67 $\pm$0.16  &  \textbf{82.15 $\pm$0.41}  & \textbf{80.79 $\pm$0.34}  &  \textbf{68.25 $\pm$0.43}  &  	\textbf{84.66  $\pm$0.23}  \\
        \vspace{0.05cm}
        AAPA-10k 	& 88.75 $\pm$0.15  &  \textbf{82.15 $\pm$0.41}  & 80.64 $\pm$0.35  &  68.22 $\pm$0.44  &  	84.63  $\pm$0.24  \\
        \textbf{b) m$L^2$} $\pm$SEM &  &  &  &  &  \\ %
        \hline
        OPNet     & {1.42 $\pm$0.07} & 3.44 $\pm$0.23 & 7.33 $\pm$0.41 & 13.97 $\pm$0.51 & 4.53 $\pm$0.26 \\
        OPNet+AA & 1.45 $\pm$0.06 & 	\textbf{2.41 $\pm$0.11} & 	7.98 $\pm$0.41 & 	16.68 $\pm$0.51 & 	4.56 $\pm$0.24 \\ 
        PA-6k5 & $3.21\pm0.28$ & 5.19 $\pm$0.42 & 13.04 $\pm$0.55 & 29.92 $\pm$0.53 & 7.93 $\pm$0.39 \\ 
        AAPA-3k & 1.67  $\pm$0.19 & 6.27  $\pm$1.11 & 3.05  $\pm$0.47 & 4.67  $\pm$0.37 & 3.09  $\pm$0.44 \\
        AAPA-6k5 & 1.44 $\pm$0.09 & {2.44 $\pm$0.18} & \textbf{2.25 $\pm$0.13} & \textbf{4.65 $\pm$0.21} & \textbf{1.96 $\pm$0.11} \\
        \vspace{0.05cm}
        AAPA-10k & \textbf{1.35  $\pm$0.08} & 2.44  $\pm$1.18 & 2.64  $\pm$0.20 & 5.12  $\pm$0.28 & 2.10  $\pm$0.13 \\
        \hline
        \multicolumn{6}{|c|}{\textbf{II. Using Perfect Perception (PP)}} \\
        \hline
        \textbf{c) mIoU} $\pm$SEM & Visible & Occluded & Contained & Carried & Overall \\
        \hline
        OPNet 	    & 90.02 $\pm$0.11  &	73.81 $\pm$0.57  & 	84.40 $\pm$0.42  & 	77.08 $\pm$0.64  & 	87.09 $\pm$0.28\\
        OPNet+AA  	& 93.23 $\pm$0.07  &	87.91 $\pm$0.50  & 	89.97 $\pm$0.15  & 	\textbf{87.81 $\pm$0.21}  & 	91.74 $\pm$0.11\\ 
        AAPA-6k5	    & 99.38 $\pm$0.12  &	\textbf{90.53 $\pm$0.64}  & 	93.86 $\pm$0.33  & 	82.54 $\pm$0.34  & 	96.31 $\pm$0.22\\
        \vspace{0.05cm}
        AAPA-10k 	& \textbf{99.60 $\pm$0.06}  &	\textbf{90.53 $\pm$0.64}  & 	\textbf{94.16 $\pm$0.30}  & 	83.02 $\pm$0.31  & 	\textbf{96.54 $\pm$0.19}\\
        
        \textbf{d) m$L^2$} $\pm$SEM &  &  &  &  &  \\ %
        \hline
        OPNet     & 1.07 $\pm$0.04 & 4.16 $\pm$0.17 &  3.44 $\pm$0.28 & 5.75 $\pm$0.34 & 2.28 $\pm$0.17 \\
        OPNet+AA & 0.69 $\pm$0.02 & \textbf{2.12 $\pm$0.15} & 1.07 $\pm$0.02 & \textbf{1.42 $\pm$0.04} & 0.92 $\pm$0.04\\ 
        AAPA-6k5 & 0.28 $\pm$0.08 & 2.51 $\pm$0.23 &  0.77 $\pm$0.15 & 1.94 $\pm$0.17 & 0.62 $\pm$0.11 \\
        AAPA-10k & \textbf{0.11 $\pm$0.03} & 2.51 $\pm$0.23 & \textbf{0.46 $\pm$0.06} & 1.48 $\pm$0.07 & \textbf{0.42 $\pm$0.05} \\ 
        \end{tabular}
    };
     
  \end{tikzpicture}
    	\vskip -0.2in
\end{table*}


\subsection{Results and analyzes}
\label{subsec:quant-results}
In the following we first compare overall model performances using object detection (OD) and perfect perception (PP), by focusing on carried tasks. Then, we discuss the performance of models with and without action annotations and different model variants. We further analyze both models that use action annotations (OPNet+AA vs AAPA) and conclude with a discussion on common failure cases.

\textbf{1) Overall model performances using OD and PP:} \label{subsec:baseline-comp}
\newline
A comparison of model performances are shown in Table~\ref{tab:model-comparison}.I using OD and Table~\ref{tab:model-comparison}.II using PP.
Since the videos are synthetically rendered, there was considerably little noise (compared to a real camera feed) and the Faster R-CNN produced few detection errors.
Nevertheless, since there were no detection errors at all using PP, overall model performances were better.

Using OD, AAPA-6k5 and AAPA-10k performed best overall, with accuracy of 84.6\%, exceeding other performances by more than 2\%.  AAPA-6k5 significantly outperforms other models for carried tasks when using mean IoU (AAPA-6k5 68.25\% vs OPNet 56.04\%) as well as mean $L^2$-distance (AAPA-6k5 4.65px vs OPNet 13.97px).

Using PP, AAPA-10k performed best overall, with an accuracy of 96.54\%, exceeding other performances by more than 4\%. However, OPNet+AA performed best for carried tasks when using mean IoU (OPNet+AA 87.81\% vs AAPA-10k 83.02\%) as well as mean $L^2$-distance (OPNet+AA 1.42px vs AAPA-10k 1.48px).

For AAPA models, main failure cases were due to tracking failures, as the snitch moves while in occlusion, as well as bounding box prediction errors.
Therefore, failure cases for carried tasks were caused by the snitch being contained, followed by the tracked object moving in occlusion.
Other failures were due to anchoring errors, where the object alignment threshold $\tau$ was too low (not allowing big object leaps in two consecutive frames) or too high (giving more leeway to detection errors).


\textbf{2) Models with and without action annotations:}\label{subsec:models-aa}
\newline
\textit{{a) OPNet vs OPNet+AA:}} When using OD, OPNet+AA performs equally or slightly worse than OPNet. As the detection module occasionally misclassifies objects in some frames, objects that seem to be visible in the video are considered missing by the reasoning model causing the flickering errors.
This might explain why our action vector in OPNet+AA did not increase OPNet's performance by a lot: in some frames, even if there is a correct weight, the detection module does not detect the parent object containing the snitch and classified it as something else. As a result, the bounding box information of the parent object is a zero vector (\ie [0,0,0,0,0,0]) and the product will be 0.
When using Perfect Perception, these detection errors do not exist, and the performance of OPNet+AA improves significantly, outperforming OPNet by 10.7\% on carried tasks and 4.6\% overall.
\newline
\textit{b) {PA vs AAPA:}} 
PA-6k5, the perceptual anchoring model without object attachments, performed a lot worse on all tasks compared to AAPA-6k5, its equivalent with action annotations. The performance on carried tasks drops drastically as expected (since it does not follow the parent object at all). 
All AAPA models perform significantly better than PA, showing the impact of using agent actions and attachment relations.

\textbf{3) AAPA model variant comparison:}\label{subsec:aapa-variants}
\newline
When setting the object alignment threshold to $\tau = 3000$, object state changes in two consecutive frames were stricter, meaning that the model relies less on the object detector output.
In cases where flickering errors occurred that detected the snitch at an arbitrary location, AAPA-3k maintained the correct anchor, due to the assumption that objects do not teleport or take big leaps. However, this lead to failure cases, when the snitch moved bigger distances between two frames. Overall, AAPA-3k performed worse than OPNet and other AAPA variants with higher thresholds. The performance of AAPA-3k would likely decrease using PP but improve with more noisy object detectors.

Setting $\tau = 6500$ yielded the best overall results for AAPA using OD, as it provides a good middle ground to handle a lot of the flickering detection errors but also captures objects that take slightly bigger leaps between consecutive frames. Using PP, there were no detection errors, thus AAPA-6k5 is outperformed by AAPA-10k which captures all objects that take bigger leaps. 

With $\tau = 10000$, AAPA-10k is more lenient on object alignment dissimilarities, thus `trusting' the output of the object detector, but giving more leeway for objects that take bigger leaps but also for flickering errors. Using OD, most failure cases occurred due to flickering errors, where the snitch was detected in another position. While it performs better on cases where objects took bigger leaps, it does not perform as well in comparison to AAPA-6k5. 
As expected, the performance of AAPA-10k improves significantly when using PP, outperforming AAPA-6k5 overall and on carried tasks. This is due to the fact, that there are no flickering errors and all object leaps are anchored correctly.

\textbf{4) OPNet+AA vs AAPA:}\newline
As discussed previously, OPNet+AA did not perform well when using OD, therefore AAPA-6k5 outperforms the neural model in all cases.
However, using PP, OPNet+AA improved the most on carried tasks (OPNet+AA 87.81\% vs AAPA-10k 83.02\%), outperforming all other models.
Note that when comparing $L^2$-distances, the performances of both AAPA-10k and OPNet+AA are very similar, with a maximum difference of 0.6 pixel for all tasks, which can be considered insignificant. 
The inferior performance of AAPA-10k by 4\% using mean IoU can be attributed to incorrect bounding box size predictions: As AAPA maintains the last detected bounding box size before it is contained and carried, the predicted size is wrong when it moves in space. OPNet+AA learns to adjust the size in the training phase.

\subsection{Discussions}\label{subsec:discussions}
\begin{figure}[h]
	\centering	
 	\includegraphics[width=0.99\linewidth]{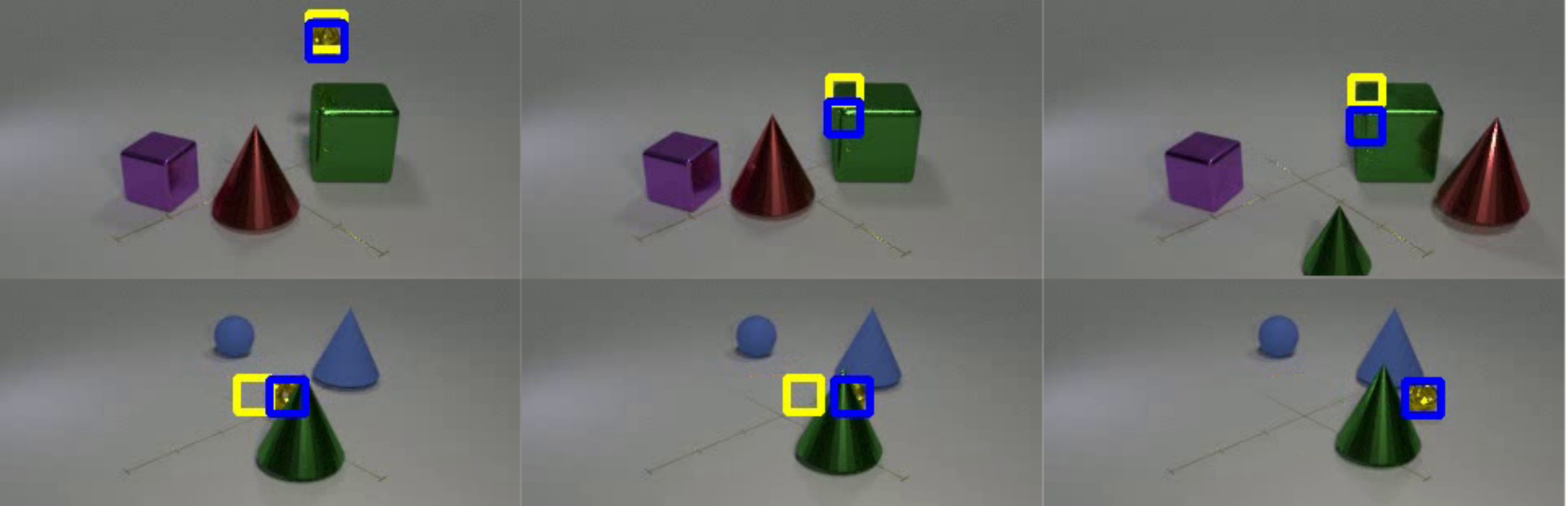}%
 	\vskip -0.1in
 	\caption{Example of common tracking failures when the object is moving under occlusion: AAPA fails to track it if it stops (top), if it reappears, AAPA anchors it correctly again (bottom).}
	\label{fig:place-in-occl}
	\vskip -0.1in
\end{figure}

With these experiments we showed the potential applicability and benefits of using action annotation for object permanence in both neural and rule-based models. It is important to remember, however, that the neural models need a large amount of data for training and validation and their failure cases are not inherently explainable. For OPNet+AA, the main failures occurred due to the OD not detecting the parent object, which was confirmed by its outstanding performance using PP. While we tried different weight parameters with and without SoftMax for OPNet+AA, it was difficult for us to point out other causes for failures. AAPA's rule-based nature allows us to analyze causes for tracking errors and to fine-tune the parameters of our model without requiring any training.
We conclude this section by addressing the main failure cases observed for AAPA:

\textbf{Moving under occlusion:} 
As we assume that objects do not move on their own, AAPA fails to track the snitch when it was moving while under occlusion. If the snitch is merely passing behind an object and reappears within a couple of frames, it is still anchored it correctly. 
However, if the snitch stops behind the object, AAPA tracks the last position before it became occluded (usually at the border of the occluder), therefore affecting both mean IoU and $L^2$-distance (Fig.~\ref{fig:place-in-occl}). 
This also affects the performance on the contained and carried tasks, if the snitch is being contained after moving behind an object.

\textbf{Bounding box misalignment:}
Since AAPA retains the last detected bounding box size, the mean IoU is less accurate when the object is carried in 3D space as the ground truth size becomes smaller or larger.
The neural models (OPNet and OPNet+AA) both learn to adjust the bounding box size during training. This explains why OPNet+AA performs best for carried tasks using PP, as it benefits from both attachment relations and bounding box size adjustment.
As these differences in bounding box size predictions can impact the performance of the models, we also compared $L^2$-distances to get a better estimate of how reliable the model predictions are, \eg in Table~\ref{tab:model-comparison}.I for visible tasks, OPNet performs best using mean IoU, but not using $L^2$-distances. 
\vspace{-0.05in}

\section{SYSTEM EVALUATION}\label{sec:sys-eval}
\vspace{-0.05in}
To demonstrate the impact of using agent actions in a real robotic application, we perform a system evaluation of AAPA's object permanence capabilities.
We evaluate the model on benchmark tasks in two scenarios (Tab.~\ref{tab:task_list}):
\begin{enumerate}[leftmargin=*,noitemsep]
    \item with a Universal Robot UR5 cobot, an industrial robotic arm with 6 DoF and a load capacity of 5kg placed in a lab setting with 3D-printed gearbox parts (Fig.~\ref{fig:demo-overview}), and
    \item with a Universal Robot UR16e cobot with 6 DoF and a load capacity of 16kg placed in an industrial manufacturing environment with real metal gearbox parts (Fig.~\ref{fig:demo2}).
\end{enumerate}

Both robots were mounted with an Intel RealSense D435 camera for an eye-in-hand system. 
We used the cognitive architecture \icarus/~\cite{Choi2018} to integrate AAPA and the communication with ROS. \icarus/ includes functionalities for world state inference (from a given world model) and high-level task planning.
To control the robots we used the python-urx library\footnote{{https://github.com/SintefManufacturing/python-urx}} and KOVIS \cite{puang2020kovis}, a Keypoint-based Visual Servoing method for fine object manipulation tasks such as grasping and insertion. 
We used a Faster R-CNN, trained to detect objects for our gearbox assembly task (such as gearbox case, input/output subassemblies, hub cover \etc). 

\begin{table}[thbp]
\caption{Benchmark tasks used for the system evaluation}
	\vskip -0.2in
\label{tab:task_list}
\begin{center}
\begin{tabular}{cll}
\# & Task & Robot \\ \hline
1.1 & Attach hub cover to case & UR5 \\
1.2 & Insert input subassembly to case & UR5 \\
1.3 & Assemble gearbox top \& bottom (collaboration) & UR5 \\
1.4 & Screw plug into case & UR5 \\
2.1 & Insert output subassembly to case & UR16e \\
2.2 & Insert three subassemblies to case (collaboration) & UR16e 
\end{tabular}
\end{center}
\vskip -0.2in
\end{table}

\subsection{Experimental setup \& scenarios}
\begin{figure}[h]
	\centering	
 	\includegraphics[width=1\linewidth]{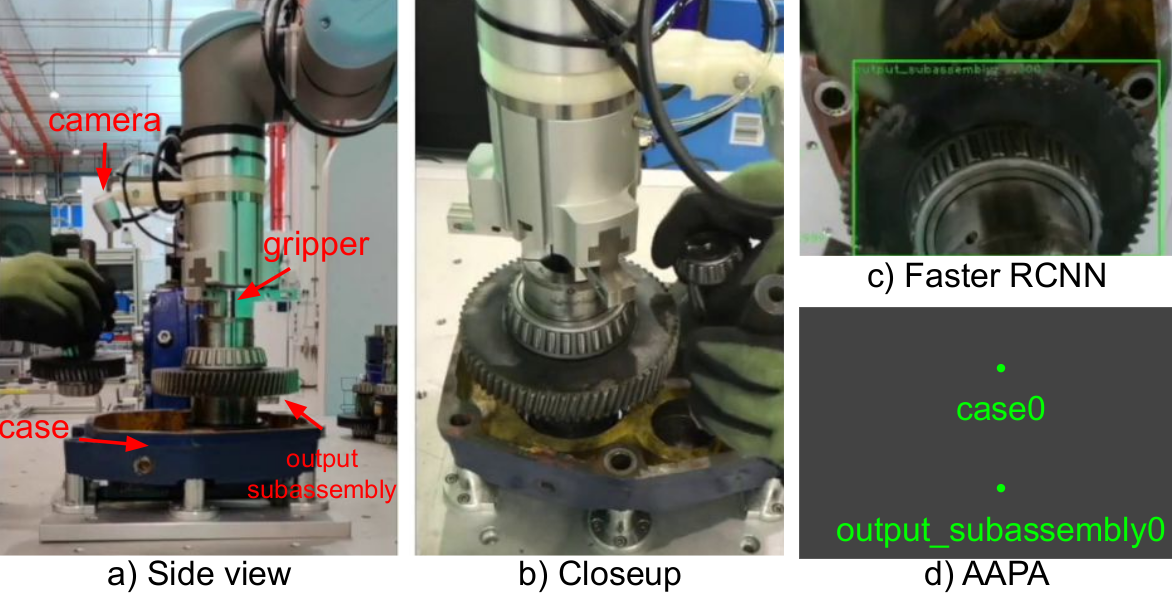}%
 	\vskip -0.1in
 	\caption{Collaborative task in an industrial environment (Task 2.2): the robot's camera view is obstructed by the output subassembly, the object detector is not able to detect the case using the Faster RCNN (c), but it is still maintained by AAPA (d).}
	\label{fig:demo2}
	\vskip -0.1in
\end{figure}

The gearbox assembly consists of different subtasks, which can either be executed by the robot alone (Tasks \#1.1, 1.2, 1.4, 2.1) or require a human-robot collaboration (\#1.3, 2.2) as heavy parts need to be assembled simultaneously.
Task 1.3 (Fig.~\ref{fig:demo-overview}) requires the robot to pick and place the casing top on the casing base which is held by the human operator.
Task 2.2 (Fig.~\ref{fig:demo2}) requires the robot to pick up the output subassembly in order to insert it with the other subassemblies manipulated by the human operator.
The robot uses AAPA to maintain the world model of the detected objects, from which \icarus/ infers new object relations and defines the next executable action to achieve the goal. 
We created a gearbox assembly domain in \icarus/, where actions, objects and their relations are defined as conceptual rules or structures with attributes. For example, the robot's pick-up action is defined as: 
\begin{verbatim}
((pick-up ?hand ?obj)
 :elements ((hand ?hand)
         (object ?obj *type output)
         (in-reachable-pose ?hand ?obj)
         (not (holding ?hand ?any)))
 :actions ((*grasp ?hand ?obj))
 :effects ((holding ?hand ?obj)))
\end{verbatim}
Similarly, for actions such as {insert} and {screw-in}.
For AAPA to use these actions in the hypothesis reasoning process with the attachment hierarchy, we used $\mathcal{AD}=\{$(pick-up,place-down), (screw-in,unscrew), (insert,take-out)\}. 
We first adjusted the model parameters for AAPA such as the object alignment threshold $\tau$ to account for noisy data and lighting conditions in different settings. The same parameter settings were used for all experimental tasks.
For each task, we set a corresponding concept goal for the robot to achieve. 
We ran the task executions with and without considering agent actions, \ie by removing the attachment concepts.\footnote{Sample executions can be seen at {https://iros2021.page.link/video}} 

\vspace{-0.1in}
\begin{figure}[h]
	\centering	
 	\includegraphics[width=1\linewidth]{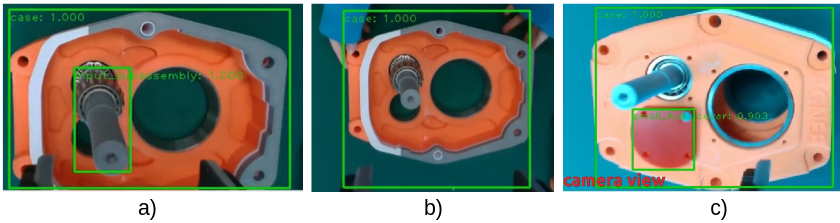}%
 	\vskip -0.1in
 	\caption{Example of detection failures due to noise and lighting conditions: the input subassembly is still detected after inserting into the case (a), but not from a different angle (b) or when it is partially occluded (c).}
	\label{fig:detection-loss}
	\vskip -0.1in
\end{figure}

\vskip -0.1in
\subsection{Results}
When running the model without the attachment relations, the task execution would fail when an object was not detected anymore from being grasped (\ie due to the obstructed camera view or viewing angle), from being attached to the parent object (\ie due to the cluttered environment), or simply due to different lighting conditions (Fig.~\ref{fig:detection-loss}).
With the attachment relations being inferred from agent actions, the robots were able to keep track of lost objects.
Adjusting the AAPA parameters at the start was necessary as, unlike the LA-CATER dataset, where we used synthetically rendered videos, the camera input to the object detector contained a lot more noise (\eg due to different lighting conditions) and caused flickering effects of detected bounding boxes. There were also multiple objects of the same type, which caused detection errors and unique ID switches by the Faster R-CNN.
However, using AAPA, both robots managed to anchor lost objects and completed all tasks successfully.

\vspace{-0.05in}
\section{Conclusions}\label{sec:conclusion}
\vspace{-0.05in}
In this paper, we argue for the use of agent actions for improved object permanence. We proposed an approach to infer attachment relations from agent actions that can be used for rule-based models, such as AAPA~\cite{icra2021}, as well as for purely neural models, such as OPNet~\cite{shamsian2020learning}.
Our approach combines visually perceived information with agent actions to keep track of objects for more complicated tasks such as obscured, carried objects.
We compared performances of different object permanence models, where those that use agent actions significantly outperformed others on a snitch localization task using the LA-CATER dataset of 1.4K videos.
We further demonstrate the usability of our approach in real-world applications with qualitative experiments on two Universal Robots (UR5 and UR16e) in both lab and industrial settings, where our robotic eye-in-hand system uses AAPA to infer hidden object states from agent actions.

We plan to further investigate the use of actions in relation with object state changes, as well as for several improvements to AAPA. 
While we only considered few actions in the experiments, attachments can be inferred from others, such as {fasten, nail}, as well as from other domains like multi-object tracking, where humans enter vehicles and therefore create an attachment relation. We will expand our library of such actions to address different use cases and exploit the generality of our reasoning process further.
Furthermore, we currently defined these attach-detach actions manually and used given action annotations or the robot's own executed actions. To infer general attachment relations in a collaborative setting, the robot needs to be able to detect human actions and learn attach-detach actions from observation. 

\addtolength{\textheight}{-12cm}   


\vspace{-0.1in}
\small{
\section*{Acknowledgements}
\vspace{-0.05in}
This research is supported by A$^{*}$STAR under its Human-Robot Collaborative AI for Advanced Manufacturing and Engineering (AME) programme (Grant number A18A2b0046).
}
\bibliographystyle{ieee_fullname}
\bibliography{egbib}

\end{document}